\pdfoutput=1

\documentclass[11pt]{article}

\usepackage[]{acl}

\usepackage{times}
\usepackage{latexsym}

\usepackage[T1]{fontenc}

\usepackage[utf8]{inputenc}

\usepackage{microtype}

\usepackage{inconsolata}

\usepackage{amsmath}
\usepackage{amsfonts}
\usepackage{graphicx}
\usepackage{amssymb}
\usepackage{booktabs} 
\usepackage{musicography}
\usepackage{multirow}
\usepackage[inline]{enumitem} 
\usepackage{graphicx}
\usepackage{subfig}
\usepackage{xcolor}
\definecolor{LightSteelBlue1}{RGB}{202,225,255}
\definecolor{Moccasin}{RGB}{255, 228, 181}
\usepackage{tikz}
\usepackage{pifont}
\title{GraSAME: Injecting Token-Level Structural Information to Pretrained Language Models via Graph-guided Self-Attention Mechanism}

\author{Shuzhou Yuan \and Michael Färber \\
        Karlsruhe Institute of Technology \\ 
         Center for Scalable Data Analytics and 
Artificial Intelligence (ScaDS.AI)\\
        TU Dresden \\ 
        \texttt{\{shuzhou.yuan, michael.faerber\}@kit.edu}}

\begin{document}
\maketitle
\begin{abstract}
Pretrained Language Models (PLMs) benefit from external knowledge stored in graph structures for various downstream tasks. However, bridging the modality gap between graph structures and text remains a significant challenge. Traditional methods like linearizing graphs for PLMs lose vital graph connectivity, whereas Graph Neural Networks (GNNs) require cumbersome processes 
for integration into PLMs. In this work, we propose a novel graph-guided self-attention mechanism, GraSAME. GraSAME seamlessly incorporates token-level structural information into PLMs without necessitating additional alignment or concatenation efforts. As an end-to-end, lightweight multimodal module, GraSAME follows a multi-task learning strategy and effectively bridges the gap between graph and textual modalities, facilitating dynamic interactions between GNNs and PLMs. Our experiments on the graph-to-text generation task demonstrate that GraSAME outperforms baseline models and achieves results comparable to state-of-the-art (SOTA) models on WebNLG datasets. Furthermore, compared to SOTA models, GraSAME eliminates the need for extra pre-training tasks to adjust graph inputs and reduces the number of trainable parameters by over 100 million.

\end{abstract}

\section{Introduction}
	\begin{figure*}[htbp]
    \centering
    \includegraphics[width=1\textwidth]{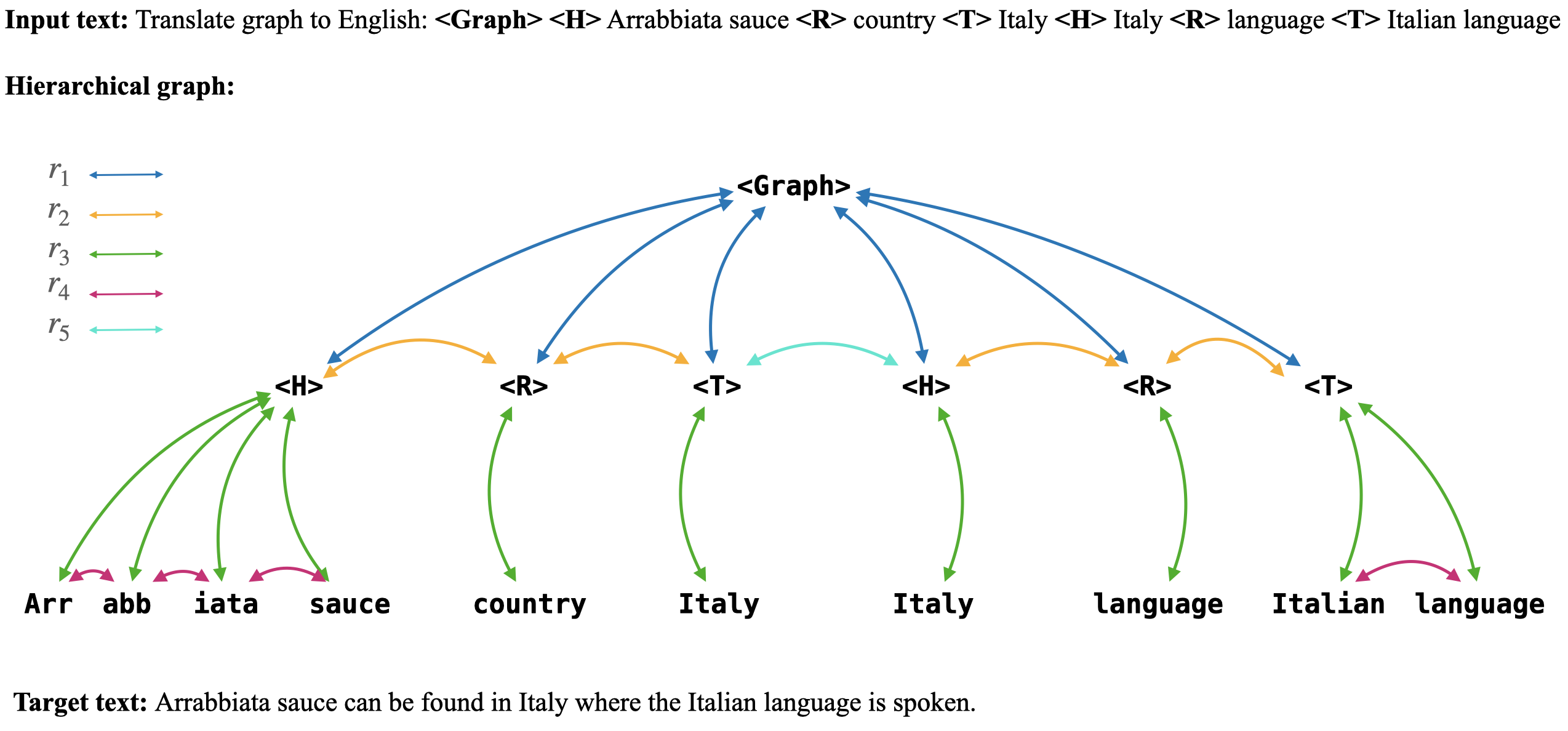}
    \caption{An example of KG-to-text generation. We visualize the hierarchical graph structure derived from the linearized graph input, tokenized by T5-large tokenizer. Each token from the input text is depicted as a node. Various relation types, each indicated by a unique color, are assigned between nodes to establish the hierarchical structure and ensure effective information flow among neighboring nodes.}
    \label{hierarchical_graph}
\end{figure*}

The paradigm of pre-training and fine-tuning has increasingly become the standard approach for leveraging the inherent knowledge of language models in a wide range of Natural Language Processing (NLP) tasks \citep{xu-etal-2021-rethinking}. Pretrained Language Models (PLMs) like Transformer \citep{vaswani2017attention}, T5 \citep{raffel2020exploring}, and GPT \citep{brown2020language}, which are trained on extensive text corpora, have demonstrated remarkable performance across various NLP challenges. However, these models primarily focus on textual data, presenting a significant limitation in processing structured information, such as Knowledge Graphs (KGs), molecular graph and social networks. Such graph structures are crucial for storing external knowledge and can significantly enhance PLM's performance on knowledge-driven tasks \citep{zhang-etal-2019-ernie, peters-etal-2019-knowledge}. This limitation becomes particularly evident in tasks that require a deep understanding of both textual and structural data, such as graph-to-text generation \citep{gardent-etal-2017-webnlg}, KG-based fact checking \citep{kim-etal-2023-factkg}, 
and translation between molecules and natural language \citep{edwards-etal-2022-translation}. 

To address the challenge of processing structural input in PLMs, recent research has explored two main strategies: linearizing graph structures into text sequences \citep{harkous-etal-2020-text,ribeiro-etal-2021-investigating,schmitt-etal-2021-modeling}, and encoding structural information using Graph Neural Networks (GNNs) \citep{yao-etal-2020-heterogeneous,ribeiro-etal-2021-structural,li-etal-2021-shot-knowledge,zhang2022greaselm}. While linearization allows direct fine-tuning of PLMs, studies have shown that it often fails to preserve the inherent structural information and explicit node connectivity \citep{song-etal-2018-graph,ribeiro-etal-2019-enhancing}. Conversely, while GNNs effectively encode complex structures, the modality difference between text and graphs complicates the integration with PLMs, requiring additional training for aligning and concatenating embeddings from graph and textual modality \citep{li-etal-2021-shot-knowledge,zhang2022greaselm}.

To merge the strengths of PLMs and GNNs, we introduce GraSAME, a novel \textbf{Gra}ph-guided \textbf{S}elf-\textbf{A}ttention \textbf{ME}chanism, enhancing PLMs' ability to process graph inputs. As depicted in Figure \ref{hierarchical_graph}, we construct a token-level hierarchical graph structure from the linearized graph sequence to maintain the input graph's structural integrity. GraSAME is designed to learn the token-level graph information from a GNN and integrate it seamlessly into the text representation for PLM. Based on a standard transformer architecture, we substitute the self-attention layers in the encoder with GraSAME. As GraSAME effectively encodes the graph structure, we train only its parameters while keeping the PLM's parameters frozen. This approach enables PLMs equipped with GraSAME to simultaneously process structural and textual information dynamically, eliminating the need for complex alignment or concatenation of different modalities. 

Applied to KG-to-text generation task WebNLG, we integrate GraSAME into the encoder-decoder model T5 with multi-task fine-tuning. The KG-to-text generation is particularly suitable as it necessitates the processing of both graph and text information, providing a clear intuition to assess the effectiveness of GraSAME. 
Our experiments demonstrate that GraSAME is compatible with various GNN architectures such as GraphSAGE \citep{hamilton2017inductive}, GAT \citep{velickovic2018graph} and RGCN \citep{schlichtkrull2018modeling}, yielding performance that not only surpasses baseline models but also comparable to state-of-the-art (SOTA) models. Moreover, GraSAME effectively integrates structural information purely during fine-tuning and saves over 100 million trainable parameters. 


In summary, our contributions are: 
\begin{enumerate*}[label={\textbf{\roman{*})}}]
		\item Introducing a novel graph-guided attention mechanism GraSAME to incorporate explicit structural information into PLMs. 
  This innovation enables PLMs to process both textual and structural inputs smoothly, bridging the modality gap of GNNs and PLMs. With GraSAME, PLMs can dynamically interact with GNNs, effectively interpreting graph inputs, which is crucial for NLP tasks that require structural information.
		\item Applying GraSAME to KG-to-text generation on WebNLG datasets, achieving results comparable to SOTA models while saving over 100 million trainable parameters.
	\end{enumerate*}

\section{Related Work}
\textbf{Structural Information for PLMs.} Although PLMs inherit linguistic structure information from pre-training \citep{nie2024decomposed}, external structural information helps PLMs enhance their ability to understand the syntax of natural language \citep{yang-etal-2022-learning-structural}, summarize source code \citep{choi-etal-2021-learning} and generate better text \citep{song-etal-2020-structural}. Much initial focus of infusing structural information into PLMs has been on modifying pre-training objectives \citep{peters-etal-2019-knowledge,xiong2019pretrained,he-etal-2020-bert}. \citet{zhang-etal-2019-ernie} utilized both textual corpora and KGs to pre-train an enhanced language representation model. \citet{ke-etal-2021-jointgt} proposed three new pre-training tasks to explicitly enhance the graph-text alignment. Also, recent efforts increasingly aimed at injecting structural information into PLMs during fine-tuning for various NLP tasks \citep{yasunaga-etal-2021-qa,ribeiro-etal-2021-structural}. \citet{wang-etal-2021-k} proposed K-Adapter to infuse knowledge into PLMs. \citet{zhang2022greaselm} came up with GreaseLM model to utilise KG information for question answering. 

\textbf{KG-to-text Generation.} Previous approaches to enabling PLMs to process graph inputs often relied on linearizing the input graph into a text sequence \citep{harkous-etal-2020-text,mager-etal-2020-gpt,ribeiro-etal-2021-structural,colas-etal-2022-gap}. \citet{ribeiro-etal-2021-investigating} investigated PLMs on graph-to-text generation using linearized graphs and found that this method is effective, yet results in the loss of specific edge connections in graphs \citep{song-etal-2018-graph,beck-etal-2018-graph,ribeiro-etal-2019-enhancing}. Also, \citet{yuan-faerber-2023-evaluating} evaluated generative models using linearized graph and uncovered the issues of hallucinations in the models. An alternative approach to encode the graph inputs is leveraging GNNs \citep{koncel-kedziorski-etal-2019-text,ribeiro-etal-2020-modeling}. But this method typically entails additional steps such as aligning modalities and concatenating embeddings \citep{li-etal-2021-shot-knowledge}, which adds complexity to the development of a seamless end-to-end pipeline for integrating GNN with PLM. Diverging from the previous methods, our work synthesizes the strengths of both linearized graph and GNN. Moreover, GraSAME also follows a lightweight fine-tuning avoiding updating the parameters of the whole model, inspired by the adapter and parameter-efficient fine-tuning approaches \citep{houlsby2019parameter,ribeiro-etal-2021-structural,wang-etal-2021-k, yuan2024gnnavi}.

\section{Model}
In this section, we detail the components of our model. Theoretically, GraSAME is adaptable to any attention-based PLMs. 
We choose T5 model \citep{raffel2020exploring} as our foundation due to its encoder-decoder architecture, which is well-suited for KG-to-text generation. 

\subsection{Encoder-Decoder Model}
Encoder-decoder model, such as T5, is a classic Transformer model consisting of encoder and decoder layers. Each encoder layer includes two distinct sublayers: a self-attention mechanism and a position-wise fully connected feed-forward network. The self-attention mechanism utilizes $h$ distinct attention heads. Consider a conditional generation task such as KG-to-text generation, where the input is a sequence of tokens $x = (x_1, \ldots, x_n)$ with each $x_i \in \mathbb{R}^{d_x}$, and the aim is to generate target sequence of tokens $y = (y_1, \ldots, y_n)$. The attention head processes an input sequence, the outputs of all attention heads are merged via concatenation, followed by a parameterized linear transformation to yield the final output of the self-attention sublayer. The computation of each output element $z_i$, with each $z_i \in \mathbb{R}^{d_z}$, involves a weighted sum of linearly transformed input elements, defined as:

\begin{equation}\label{eq_1}
    z_i = \sum_{j=1}^{n} \alpha_{ij} (x_j W^V),
\end{equation}
where $\alpha_{ij}$ represents the weight coefficient, calculated using a softmax function:
\begin{equation}\label{eq_2}
    \alpha_{ij} = \text{softmax}(\frac{(x_i W^Q)(x_j W^K)^T}{\sqrt{d_k}}).
\end{equation}
The matrices $W^V, W^Q, W^K\in \mathbb{R}^{d_x \times d_z}$ are layer-specific trainable parameters, and are distinct for each attention head.

\subsection{Graph-guided Self-Attention Mechanism}
Self-attention allows for the interaction of token representations by treating each input sequence as a fully-connected graph with tokens as nodes \citep{yao-wan-2020-multimodal}. However, this process does not retain the original structural information and explicit connectivity between the tokens. To address this issue, we introduce GraSAME, a method that integrates text with token-level hierarchical graph representation illustrated in Figure \ref{hierarchical_graph}.
\begin{figure}[tbh]
    \centering
    \includegraphics[width=0.48\textwidth]{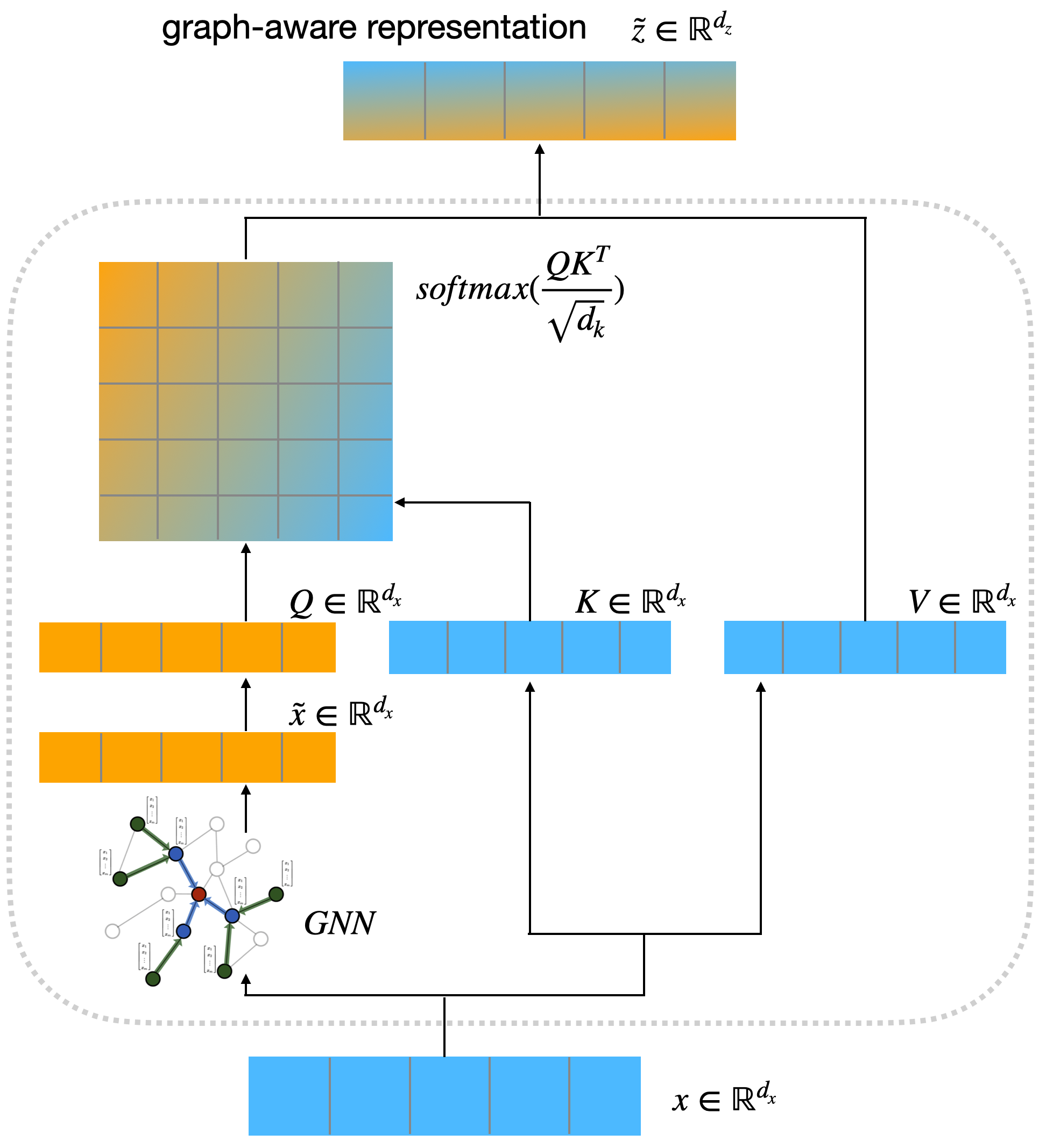}
    \caption{The architecture of GraSAME. The \colorbox{LightSteelBlue1}{text embedding $x$} is fed into a GNN along with the edge index derived from the hierarchical graph structure. This process generates a \colorbox{Moccasin}{graph embedding $\tilde{x}$}, which subsequently induces the $Q$ vector. The $Q$ vector then guides the self-attention mechanism to produce a
    graph-aware representation $\tilde{z}$. The visualization of GNN is taken from GraphSAGE \citep{hamilton2017inductive}.}
    \label{attention_compute}
\end{figure}
\subsubsection{Architecture of GraSAME}
GraSAME involves incorporating a GNN within the self-attention layer of the PLM. This addition enables the direct encoding of the hierarchical graph structure and facilitates the smooth transfer of structural information into the PLM. We visualize the architecture in Figure \ref{attention_compute}. 

\textbf{Graph Neural Network.} The graph neural network models structural characteristics of the input graph effectively by using various graph convolutional layers.
The primary goal of the GNN is to learn the representations for both individual nodes and the overall graph structure. In most GNN models, the node representation is updated iteratively by aggregating the representations of its neighboring nodes. The representation of node $v_i$ at the $l$-th layer is represented by $h^{(l)}$, with the initial representation $h^{(0)}$ set to the node's feature vector $x_i$. The process of representation update at the $l$-th layer involves two main operations:
\begin{align}
a^{(l)} &= \text{AGGREGATE}^{(l)} \left( \{ h^{(l-1)}_j : j \in \mathcal{N}(v_i) \} \right), \\
h^{(l)}_i &= \text{COMBINE}^{(l)} \left( h^{(l-1)}_i, a^{(l)} \right),
\end{align}
where $\mathcal{N}(v_i)$ denotes the neighbors of $v_i$. The AGGREGATE function compiles message passing from these neighbors, using techniques like MEAN, MAX, or SUM, varying with the GNN architecture. The COMBINE function then integrates this aggregated information into the node's current representation, thereby updating it. 

\textbf{Incorporating Method.} Drawing inspiration from the multimodal self-attention layer by \citet{yao-wan-2020-multimodal}, we present a graph-guided self-attention architecture designed to simultaneously encode text representation and hierarchical graph structure. In our approach, all tokens in the input text are treated as nodes, with their initial features derived from the token representations in $h^{(l)}$. As shown in Figure \ref{attention_compute}, the token representations are aggregated and updated through a GNN layer. This process generates the vector $Q$, which subsequently guides the self-attention layer in the encoder of the PLM.

Formally, we adapt Equation \ref{eq_2} such that the weight coefficient $\tilde{\alpha_{ij}}$ is derived from the node representation $\tilde{x_i}$ in the graph modality, and the token representation $x_j$ from the text modality:
\begin{equation}\label{eq_6}
\tilde{\alpha_{ij}} = \text{softmax}\left(\frac{(\tilde{x_i} W^Q)(x_j W^K)^T}{\sqrt{d_k}}\right).
\end{equation}
The output of the self-attention layer is then calculated as:
\begin{equation}
\tilde{z_i} = \sum_{j=1}^{n} \tilde{\alpha_{ij}} (x_j W^V).
\end{equation}
This modification ensures that the hidden word representations are influenced by the graph embedding. In each encoder layer of the model, we incorporate residual connections and layer normalization. The standard self-attention layer in the encoder is replaced with GraSAME, while the decoder retains the standard Transformer implementation. In the encoder's final layer, $\tilde{z_i}$ serves as the input to the decoder, which generates the target sequence.

\subsubsection{Graph Representation}
As PLMs are designed to process textual input only, it becomes necessary to perform certain preprocessing steps when addressing graph-based NLP tasks. Considering the task of KG-to-text generation, we represent the graph input as a linearized graph following prior studies \citep{harkous-etal-2020-text,ribeiro-etal-2021-investigating}, and also extract a token-level hierarchical graph structure to ensure information flow among neighboring nodes.

\textbf{Linearized Graph.} In line with \citet{ribeiro-etal-2021-investigating}, we linearize the graph into a sequence of text augmented with special tokens. As depicted in Figure \ref{hierarchical_graph}, a KG triple is composed of a head, relation, and tail entity. Accordingly, we prepend each entity with special tokens: <$H$>, <$R$>, <$T$>. Furthermore, to distinguish between text and graph inputs, we introduce the <$Graph$> token.\footnote{For the KG-to-text generation task, the prompt ``translate graph to English: '' is added as a task description to match the input format of T5.} Previous work \citep{ribeiro-etal-2021-investigating} suggests that PLMs generate fluent text regardless of the linearization order of the graph. Hence we adhere to the default sequence in which triples appear in the dataset.

\textbf{Hierarchical Graph Structure.} We derive a token-level hierarchical graph structure from the linearized graph, as depicted in Figure \ref{hierarchical_graph}. The concept of this hierarchy is inspired by the tree-like structures observed in functional organizations such as companies, universities, and even insect societies \citep{pooler2017hierarchical,anderson2001complexity}, and has been explored in neural network research for knowledge representation as well \citep{ying2018hierarchical,moutsinas2021graph,chen-etal-2020-kgpt}. In this structure, each token dynamically interacts within its hierarchical tier through message passing, which we believe helps preserve the graph's original explicit connectivity and ensures effective information flow among neighboring nodes. Formally, let's consider a graph \( \mathcal{G} = (\mathcal{V}, \mathcal{E}, \mathcal{R}) \), where \( \mathcal{V} \) is the set of nodes and \( \mathcal{E} \) comprises labeled edges of the form \( (u, r, v) \). Here, \( u \) and \( v \) represent nodes in \( \mathcal{V} \), and \( r \) from \( \mathcal{R} \) denotes the relation type. In our structure, each token in the linearized graph is considered as a node, and we define specific relation types \( r \) between two nodes to enhance the hierarchy:
\begin{enumerate}
    \item \( r_1 \) connects the global node <$Graph$> to special tokens <$H$>, <$R$>, <$T$>.
    \item \( r_2 \) links adjacent special tokens within a triple.
    \item \( r_3 \) connects special tokens to their respective entity tokens.
    \item \( r_4 \) joins consecutive tokens within entities.
    \item \( r_5 \) associates special tokens sharing the same entity.
\end{enumerate}
An example of such a hierarchical graph is presented in Figure \ref{hierarchical_graph}. We design the edges to be bidirectional, as this approach of information propagation in multiple directions can enhance the model's performance \citep{yao-etal-2020-heterogeneous}.

\section{Multi-Task Fine-Tuning}
Our training approach employs a multi-task learning strategy with efficient, lightweight fine-tuning. We initialize the model with pretrained parameters denoted by $\phi$. The parameters of the PLM remain frozen, and only the GNN component within GraSAME is updated, given its effective encoding of graph structure.

\subsection{Training Objectives}
We retain the standard language model objective of generating the next token in a sequence while introducing an additional graph reconstruction task. This task is designed to strengthen the relation types between pairs of nodes, enhancing the hierarchy. 

\textbf{Text Generation.} The text generation task is implemented by a PLM with a language modeling head on top. Given an input sequence $x$ and a graph representation $\mathcal{G}$, the model aims to generate a target sequence $y$ by minimizing the cross-entropy loss:
\begin{align}
    \mathcal{L}_{TG} = -\sum_{i=1}^{|y|} \log P_\phi(y_i | y_{1:i-1}, x, \mathcal{G}),
\end{align}
where $P_\phi$ is the generative probability from PLM.

\textbf{Graph Reconstruction.} Building on previous work that focused on predicting relationships between entities \citep{song-etal-2020-structural, li-etal-2021-shot-knowledge}, our approach reformulates the graph reconstruction task. We aim to predict the relation type $r$ in the triple \( (u, r, v) \), where $u$ and $v$ are nodes in the hierarchical graph structure. Node representations $h_u$ and $h_v$ are derived from the last hidden states of the PLM's encoder. Consequently, the probability of relation $r$ is given by:
\begin{align}
p(r|u, v) = \text{softmax}(W[h_u; h_v] + b),
\end{align}
where $W$ and $b$ are trainable parameters. The loss for graph reconstruction is computed using cross-entropy loss:
\begin{align}
    \mathcal{L}_{GR} = -\sum_{\langle u,r,v \rangle \in  \mathcal{E}} \log p(r|u, v).
\end{align}

We integrate the text generation loss and the graph reconstruction loss to train the PLM. The overall training loss is defined as follows:
\begin{align}\label{eq_10}
\mathcal{L}_{total} = \mathcal{L}_{TG} + \lambda\mathcal{L}_{GR},
\end{align}
where $\lambda$ is a weighting coefficient.\footnote{We observed that the value of $\lambda$ significantly impacts performance. After tuning on the validation set of the \textit{WebNLG unconstrained} dataset, we set $\lambda$ to 0.08, which yielded the best BLEU score. Further details are provided in Appendix \ref{alpha}.}

\section{Experiments}
In this section, we introduce the details of our experiments on KG-to-text generation task. We modify T5 for conditional generation from Huggingface \citep{wolf2019huggingface}, and implement GraSAME with the GNN layers provided by PyTorch Geometric \citep{Fey/Lenssen/2019}.

\subsection{Dataset}
WebNLG\footnote{\url{https://synalp.gitlabpages.inria.fr/webnlg-challenge/}} \citep{gardent-etal-2017-webnlg} is a commonly-used benchmark in KG-to-text generation \citep{chen-etal-2020-kgpt,li-etal-2021-shot-knowledge,li-liang-2021-prefix,colas-etal-2022-gap,ke-etal-2021-jointgt}. 
We employ WebNLG version 2.1 for our experiments, as it represents a refined version of the widely-used version 2.0 \citep{shimorina-gardent-2018-handling,chen-etal-2020-kgpt,ke-etal-2021-jointgt,colas-etal-2022-gap}. This version offers two distinct splits: \textit{unconstrained} and \textit{constrained}. 

\textbf{WebNLG unconstrained.} Each WebNLG sample comprises several KG triples and a corresponding descriptive text. The triples are structured as \textit{(head, relation, tail)} and the model is supposed to generate fluent text to describe the input triples.
An illustrative example is presented in Figure \ref{hierarchical_graph}. In the \textit{unconstrained} dataset, there is no overlap of input graphs between the train, validation, and test sets. 

\textbf{WebNLG constrained.} The data structure of \textit{constrained} is as same as \textit{unconstrained} dataset. However, the \textit{constrained} dataset presents a greater challenge by ensuring that there is no overlap of triples in the input graphs across train, validation and test sets. 

\subsection{Setting}
As an enhancement of self-attention mechanism, a foundational model is required for the implementation of GraSAME. For our experiments, T5-large serves as the foundational model. Prior to training, we expand T5's vocabulary to include the special tokens <$H$>, <$R$>, <$T$>, and <$Graph$>. To ensure fair comparisons, we maintain consistent hyper-parameters across both the baseline and our models.\footnote{Details on the hyper-parameters are provided in Appendix \ref{hyperparameter}.} All models are fine-tuned with the training set. The BLEU score on validation set is employed to identify the best-performing model, which is subsequently evaluated on the test set.

\textbf{Evaluation Metrics.} To evaluate the performance of the models, we use the automatic evaluation metrics BLEU \citep{papineni-etal-2002-bleu}, METEOR \citep{denkowski-lavie-2014-meteor}, ROUGE-L \citep{lin-2004-rouge} and chrF++ \citep{popovic-2015-chrf} following previous work \citep{shimorina-gardent-2018-handling,ribeiro-etal-2021-investigating}. 
The evaluations are conducted using the official evaluation script from the WebNLG challenge \citep{shimorina-gardent-2018-handling}.

\textbf{Baseline.} T5-large is employed as the baseline model. Previous research \citep{ribeiro-etal-2021-investigating,ke-etal-2021-jointgt} has demonstrated T5's SOTA performance on graph-to-text generation, making it a robust baseline \citep{clive-etal-2022-control}. This choice allows for a fair comparison with our approach.

GraSAME is integrated into the T5-large model for our experiments. We investigate the efficacy of different GNNs for encoding the hierarchical graph structure and denote them as: 

\textbf{GraSAME-GAT.} Graph Attention Network (GAT) \citep{velickovic2018graph} is used as the GNN component in GraSAME. GAT utilizes an attention mechanisms to aggregate the information from neighbouring nodes.

\textbf{GraSAME-RGCN.} We also integrate Relational Graph Convolutional Network (RGCN) \citep{schlichtkrull2018modeling} with GraSAME. RGCN extends the Graph Convolutional Network \citep{kipf2016semi}, enabling it to process local graph neighborhoods within large-scale relational data.

\textbf{GraSAME-SAGE.} GraphSAGE \citep{hamilton2017inductive} is an inductive framework that efficiently generates node embeddings for unseen nodes by leveraging node feature information. We also combine it with GraSAME.

\subsection{Evaluation Results}
\subsubsection{Main Results}
	\begin{table}[htb]   
		\centering
        \scalebox{0.7}{
		\begin{tabular}{lllllll}
  \toprule
			\textbf{Model}&\textbf{A}&\textbf{P}& \textbf{B} & \textbf{M}& \textbf{R}& \textbf{C}\\
   \midrule
            GCN & \ding{55} & - & 60.80 & 42.76 & 71.13 & - \\
            KGPT  & \checkmark &177M & 64.11 & 46.30 & 74.57 & - \\
             JointGT(T5)  &\checkmark & 265M & \underline{66.14} & \underline{47.25} & \underline{75.91} & - \\
    \midrule
             T5-large  &\ding{55} & 737M & 61.41 & 45.96  & 71.70 & 75.27 \\
			GraSAME-GAT  &\ding{55} & 75.7M & 60.44 & 44.91 & 70.73 & 72.49\\
            GraSAME-RGCN  &\ding{55} & 453M & 60.26 & 44.46 & 70.93 & 71.88\\
            GraSAME-SAGE  &\ding{55} & 151M & \textbf{65.55} & \textbf{48.38} & \textbf{74.55} & \textbf{77.34}\\
   \bottomrule
		\end{tabular}}
  \caption{\footnotesize Results on \textit{WebNLG unconstrained}. A denotes if additional pre-training tasks are implemented or not. P = Trainable parameters, B = BLEU, M = METEOR, R = ROUGE, C = chrF++. The results of GCN, KGPT, JointGT(T5) are re-printed from \citet{shimorina-gardent-2018-handling}, \citet{chen-etal-2020-kgpt} and \citet{ke-etal-2021-jointgt}, respectively. \textbf{Bold} indicates the best score of the models we trained. \underline{Underline} indicates the best score of SOTA models in previous work.
  \label{results_webnlg2u}}
	\end{table}
 
    	\begin{table*}[tb]   
		\centering
    \scalebox{1}{
		\begin{tabular}{llllllll}
  \toprule
	\textbf{Model} &\textbf{1 triple}& \textbf{2 triples} & \textbf{3 triples}& \textbf{4 triples}& \textbf{5 triples}& \textbf{6 triples}& \textbf{7 triples}\\
      \midrule
 T5 & 46.81 & 67.58 & 64.39 & 64.86 & 58.73 & 68.91 & 62.97  \\
 GraSAME  & +7.80 & +1.85 & +3.05 & +2.52 & +3.73 & -2.83 & +8.61  \\
     \bottomrule
		\end{tabular}}
  \caption{BLEU scores for different graph sizes in \textit{WebNLG unconstrained} test set. T5 is the baseline model T5-large, GraSAME is the best performed model GraSAME-SAGE. +, - denote the difference of scores.\label{results_triple}}
	\end{table*} 
We present the primary results for \textit{WebNLG unconstrained} in Table \ref{results_webnlg2u}. Remarkably, our leading model, GraSAME-SAGE, surpasses the baseline T5-large in all metrics, despite having over 500 million fewer trainable parameters. We believe this is due to the powerful encoding ability of GraphSAGE for unseen nodes (for example, the special tokens we add to the model's vocabulary manually). Although other models don't outperform the baseline T5, they still achieve noteworthy performance, achieving BLEU scores of over 60 with much fewer trainable parameters than baseline.
 
For comparison, we also include the results of the SOTA models from related work. Notably, GraSAME-SAGE outperforms GCN and KGPT on BLEU and METEOR scores, and achieves performance comparable to JoinGT(T5). Our METEOR score of 48.38 is even higher than that of JoinGT(T5). It is worth mentioning that both KGPT and JointGT are pre-trained with additional tasks to fine-tune KG-to-text generation. Additionally, JointGT(T5) has 114 million more trainable parameters than GraSAME-SAGE. This highlights the efficiency of our approach, demonstrating that GraSAME can enhance PLMs by leveraging token-level structural information for KG-to-text generation.

The results for \textit{WebNLG constrained} are similar as for \textit{WebNLG unconstrained}. As illustrated in Table \ref{results_webnlg2c}, GraSAME-SAGE outperforms the baseline T5, while our other models achieve comparable results to it with higher BLEU scores and fewer trainable parameters. In comparison to JoinGT(T5), GraSAME-SAGE maintains over 98\% performance with fewer than 114 million trainable parameters, and it doesn't require additional pre-training tasks. This prove the generalization of our method under the constrained condition, where there is no overlap between training and test triples.

 	\begin{table}[tbh]   
		\centering
    \scalebox{0.7}{
		\begin{tabular}{lllllll}
  \toprule
			\textbf{Model} &\textbf{A} &\textbf{P}& \textbf{B} & \textbf{M}& \textbf{R}& \textbf{C}\\
      \midrule
  JointGT(T5) & \checkmark & 265M & 61.01 & 46.32  & 73.57 & - \\
   \midrule
             T5-large  & \ding{55} & 737M & 58.77 & \textbf{46.12}  & 72.01 & 73.22 \\
			GraSAME-GAT & \ding{55} & 75.7M & 59.21 & 45.38 & 71.01 & 72.79\\
            GraSAME-RGCN  & \ding{55}& 453M & 60.13 & 45.47& 71.79 & 72.88\\
            GraSAME-SAGE  & \ding{55}& 151M & \textbf{60.27} & 45.81 & \textbf{72.01} & \textbf{73.29}\\
   \bottomrule
		\end{tabular}}
  \caption{\footnotesize Results on \textit{WebNLG constrained}. A denotes if additional pre-training tasks are implemented or not. P = Trainable parameters, B = BLEU, M = METEOR, R = ROUGE, C = chrF++. The results of JointGT(T5) are re-printed from \citet{ke-etal-2021-jointgt}. \textbf{Bold} indicates the best score of the models we trained.\label{results_webnlg2c}}
	\end{table}
 
\subsubsection{Detailed Analysis on Graph Size}
We conduct a comprehensive analysis focusing on input graph size, as detailed in Table \ref{results_triple}. Notably, GraSAME consistently improves the BLEU scores across various input triple counts, except for six-triple inputs. The most pronounced improvement occurs with seven-triple inputs, where the BLEU score surpasses the baseline by 8.61. Seven-triple inputs form the most complex graph structure in the test set, highlighting the efficacy of GraSAME in fortifying the PLM's capacity to process complex graph inputs through token-level structural integration. Interestingly, a significant improvement also emerges in one-triple input graphs, likely due to the baseline model's propensity for generating hallucinations with shorter input graphs.\footnote{A more in-depth error analysis is presented in Section \ref{error}. 
}

\subsection{Human Evaluation}
\begin{table}[htb]   
		\centering
		\begin{tabular}{lll}
  \toprule
			\textbf{Model} &\textbf{Fluency}& \textbf{Meaning}\\
      \midrule
  Gold  & 5.59 & 5.71 \\
  T5 & 5.57 & 5.41 \\
  GraSAME & 5.56 & 5.62 \\
   \bottomrule
		\end{tabular}
  \caption{\footnotesize Human evaluation on \textit{WebNLG unstrained}. T5 denotes the baseline model T5-large, GraSAME denotes GraSAME-SAGE. The Fleiss’ Kappa $\kappa$ is 0.42, which indicates moderate agreement.\label{human_eval}}
	\end{table}
To further assess the quality of the generated text, we conduct a human evaluation using the crowd-sourcing platform Amazon Mechanical Turk.\footnote{\url{https://www.mturk.com}} We randomly select 100 texts generated by both baseline and GraSAME-SAGE models, along with their corresponding gold standard references. In line with previous studies \citep{castro-ferreira-etal-2019-neural,ribeiro-etal-2021-investigating}, we ask three annotators to rate the texts on a 1-7 Likert scale across two dimensions:
\begin{enumerate*}[label={({\roman{*})}}]
\item Fluency: Assessing whether the text is fluent, natural, and easy to read.
\item Meaning: Evaluating if the text accurately conveys the information from the input graph without including extraneous information (hallucination).
\end{enumerate*}
We specifically instruct annotators to pay close attention to instances of hallucination, as this issue has gained significant attention in recent PLM research \citep{gonzalez-corbelle-etal-2022-dealing,ji2023survey,yuan-faerber-2023-evaluating}.

As indicated in Table \ref{human_eval}, both the baseline and GraSAME models produce fluent text, scoring only marginally lower than the reference by 0.02 and 0.03, respectively. Regarding the meaningfulness of the generated text, GraSAME surpasses the baseline, achieving a score that is 0.21 points higher. This human evaluation confirms that GraSAME is capable of generating not only fluent text but also text that more accurately encapsulates the input information, while minimizing hallucinations.

\subsection{Ablation Study}
\begin{table}[htb]   
		\centering
  \scalebox{0.90}{
		\begin{tabular}{lll}
  \toprule
			\textbf{Model} &\textbf{BLEU}& \textbf{METEOR}\\
      \midrule
  GraSAME-SAGE  & 65.55 & 48.38 \\
    \quad - bidirectional edges & 60.74 & 45.13 \\
     \quad - graph reconstruction & 61.75 & 45.65 \\
   \bottomrule
		\end{tabular}}
  \caption{Ablation study on \textit{WebNLG unconstrained}.\label{ablation}}
	\end{table}
We conduct an ablation study focusing on two key aspects of the model: the bidirectional edges and the graph reconstruction task. This study is implemented using the top-performing GraSAME-SAGE model on \textit{WebNLG unconstrained}.

\textbf{Bidirectional Edges}: We retain a single edge direction in the hierarchical graph structure, specifically from bottom to top tokens.

\textbf{Graph Reconstruction}: We omit the graph construction loss during training, allowing the model to update solely based on the text generation loss.

The outcomes of the ablation study are detailed in Table \ref{ablation}. Both the bidirectional edge and graph reconstruction components significantly enhance the performance of GraSAME. Excluding either element results in a decrease in both BLEU and METEOR scores, with a marginally greater reduction observed upon the removal of bidirectional edges. This suggests that bidirectional edges are crucial for adequate message passing within the hierarchical graph structure.

\subsection{Model Variations}
\begin{table}[htb]   
		\centering
   \scalebox{0.80}{
		\begin{tabular}{lllll}
  \toprule
			\textbf{Model} &\textbf{BLEU}& \textbf{METEOR}& \textbf{ROUGE}& \textbf{chrF++}\\
      \midrule
Variation 1 & 60.62 & 46.57 & 71.03 & 74.67\\
 Variation 2 & 60.07 & 46.01 & 70.17 & 74.08 \\
GraSAME  & 65.55 & 48.38 & 74.55 & 77.34\\
   \bottomrule
		\end{tabular}}
  \caption{Results of model variations on \textit{WebNLG unconstrained}. GraSAME = GraSAME-SAGE.\label{results_variation}}
	\end{table}
Considering the method of incorporating GNN into the self-attention layer, we introduce two additional variations of GraSAME, as visualized in Figure \ref{variation} of Appendix \ref{visual_variations}. In Variation 1, the GNN generates graph embeddings to influence the vectors $K$ and $V$, instead of $Q$. In Variation 2, we insert the GNN layer before the entire self-attention mechanism, which means the vectors $K$, $V$ and $Q$ are all derived from the graph embedding.

The results for the two variations are presented in Table \ref{results_variation}. While Variation 1 outperforms Variation 2, both variations show a decrease in performance compared to the original GraSAME. This demonstrates that our proposed GraSAME is a valid and effective mechanism for encoding structural information. 
\section{Error Analysis}\label{error}
\begin{table*}[ht]
	\centering
\footnotesize
		\begin{tabular}{lp{12cm}}
			\toprule \textbf{Model} & \textbf{Example}\\
			\midrule 
   Input & translate graph to English: <$Graph$> <$H$> Iraq <$R$> language <$T$> Arabic\\ 
   \midrule
			  Reference & Iraq language is Arabic.\\
     \midrule
            T5 & Arabic is a language spoken in the country of Iraq. \textcolor{blue}{The country is a member of the United Nations.}\\ 
            \midrule 
             GraSAME & Arabic is a language spoken in Iraq.\\
    \bottomrule
    Input & translate graph to English: <$Graph$> <$H$> Madrid <$R$> country <$T$> Spain\\ 
    \midrule
     Reference & Madrid is in the country of Spain.\\
     \midrule
            T5 & Madrid is a city in the country of Spain \textcolor{blue}{and is a popular tourist destination there.}\\ 
            \midrule 
             GraSAME & Madrid is a city in the country of Spain.\\
             \bottomrule
            Input & translate graph to English: <$Graph$> <$H$> Monocacy National Battlefield <$R$> location <$T$> Frederick County, Maryland <$H$> 14th New Jersey Volunteer Infantry Monument <$R$> established <$T$> "1907-07-11" <$H$> 14th New Jersey Volunteer Infantry Monument <$R$> country <$T$> "United States" <$H$> 14th New Jersey Volunteer Infantry Monument <$R$> category <$T$> Historic districts in the United States <$H$> 14th New Jersey Volunteer Infantry Monument <$R$> district <$T$> Monocacy National Battlefield <$H$> 14th New Jersey Volunteer Infantry Monument <$R$> state <$T$> "Maryland" \\
            \midrule
            Reference & The 14th New Jersey Volunteer Infantry Monument is located on the Monocacy National Battlefield, Frederick County, Maryland. The monument was established in 1907-07-11 and is categorised as a historic district in the United States.\\
        \midrule
            T5 & The 14th New Jersey Volunteer Infantry Monument is located \textcolor{red}{in} the Monocacy National Battlefield, Frederick County, Maryland. It was established on 11 July 1907 and is categorised as a historic district in the United States.\\ 
            \midrule 
             GraSAME & The 14th New Jersey Volunteer Infantry Monument is located on the Monocacy National Battlefield in Frederick County, Maryland, United States. It was established on 11 July 1907 and is categorised as a historic district in the United States.\\
             \bottomrule
             \end{tabular}
              \caption{Examples of text generated by T5-large and GraSAME-SAGE, with hallucinations highlighted in \textcolor{blue}{blue} and other incorrect text marked in \textcolor{red}{red}.}
			\label{error_example}
	\end{table*}
We illustrate three examples of generated text in Table \ref{error_example}. The first two examples have relatively short input texts, whereas the third example includes a longer and more complex input graph.

In the first and second examples, the text generated by T5 includes hallucinations not present in the input data. In contrast, GraSAME strictly adheres to facts conveyed by the input graphs. 
When handling more complex graph inputs, both T5 and GraSAME produce fluent and accurate text without introducing extraneous information. However, T5 makes a minor grammatical error with a preposition. This indicates that GraSAME effectively mitigates the issue of hallucinations, particularly with short and simple inputs. Moreover, for longer and more complex inputs, GraSAME demonstrates a superior understanding of the input structure, resulting in higher-quality text generation.

\section{Conclusion}
In this work, we introduce GraSAME, a novel graph-guided self-attention mechanism that enables PLMs to process token-level structural information. With GraSAME, PLMs are capable of handling text and graph input simultaneously. This approach facilitates seamless information flow between text and graph embeddings, eliminating the need for additional concatenation. 

Evaluated on the KG-to-text generation task, GraSAME demonstrates performance comparable to SOTA models with significantly fewer trainable parameters. Through a detailed analysis of graph size and human evaluation, GraSAME demonstrates its enhanced ability to process more complex graph inputs and generate more accurate text. 
Moving forward, we aim to explore GraSAME's potential in encoding specific graph structures, like molecular graph \citep{edwards-etal-2022-translation}, in combination with large language models.

\section*{Limitation}
Despite the effectiveness of our approach, we acknowledge several limitations in our work. Firstly, our method involves extracting a hierarchical graph structure from a linearized graph. While this structure facilitates efficient information exchange, it requires specific adjustments when applied to different datasets or tasks. 
Our goal is to combine the advantages of PLM and GNN, yet crafting a universal template that addresses all related tasks remains challenging.

Secondly, we observe that the training process of GraSAME is fast, but it tends to converge slower than the baseline model without GraSAME. This slower convergence is attributed to the GNN component of GraSAME not being pre-trained, necessitating additional training epochs for optimal interaction with the PLM.

Thirdly, our approach still incorporates the linearized graph as part of the input, which does not align with the pre-training process of PLMs typically conducted with plain text corpora in natural language. This misalignment could potentially lead to the forgetting of pre-trained knowledge. Addressing these limitations will be the focus of our future work.


\section*{Ethics Statement}
This research was conducted in accordance with the ACM Code of Ethics. The datasets we used are publicly available \citep{gardent-etal-2017-webnlg, kim-etal-2023-factkg}, and we only used them to evaluate our models. We are not responsible for any potentially erroneous statements within the datasets. Additionally, care should be taken with the potential issues of hallucination when using the baseline model T5 for text generation.

\section*{Acknowledgements}
We want to thank the anonymous reviewers for their helpful comments. Also, we would like to thank our colleague Yindong Wang for the inspiration of the idea, and Ercong Nie for his feedback on this work.

\bibliography{anthology,custom}


\appendix
\section{Hyperparameters}\label{hyperparameter}
We present the hyperparameters for T5-large and GraSAME in Table \ref{hyper}. We train the model until the training process converges. To keep a fair comparison, we keep the hyperparameters as same as possible. The learning rate of GraSAME is set larger than it for T5, because GraSAME converges slower due to fewer trainable parameters. The models are trained with 4 NVIDIA A100-SXM4-40GB GPUs.\footnote{We noted that when employing Distributed Data Parallel \citep{li13pytorch}, there is a possibility of sample duplication on the GPU to achieve the desired batch size, which could potentially impact the test set results. Consequently, we conducted the model evaluation on the test set using a single GPU for accuracy.}

\begin{table}[htb]   
		\centering
		\begin{tabular}{lll}
  \toprule
			\textbf{Hyperparameter} &\textbf{T5}& \textbf{GraSAME}\\
      \midrule
  \texttt{learning\_rate}  & 3e-5 & 5e-5 \\
  \texttt{optimizer}  & Adam & Adam \\
   \texttt{batch\_size} & 10 & 10 \\
    \texttt{max\_sequence\_length} & 187 & 187 \\
    \texttt{max\_target\_length}& 120 & 120 \\
    \texttt{num\_beam\_search}& 3 & 3 \\
    \texttt{random\_seed}& 123 & 123 \\
   \bottomrule
		\end{tabular}
  \caption{Hyperparameters.\label{hyper}}
	\end{table}

\section{Data Statistics}
We report the statistics of WebNLG in Table \ref{data_statis}, which is the original split in WebNLG version 2.1.
\begin{table}[ht]
  \centering
  \begin{tabular}{lccc}
    \toprule
\multirow{2}*{Dataset}   &  \multicolumn{3}{c}{Size}   \\
\cmidrule(){2-4}
       & | Train |  & | Dev |  & | Test | \\
\midrule
WebNLG U & $12876$ & $1619$ & $1600$ \\
WebNLG C & $12895$  & $1594$ & $1606$ \\

    \bottomrule
  \end{tabular}
\caption{The statistics of dataset. WebNLG U = \textit{WebNLG unconstrained}, WebNLG C = \textit{WebNLG constrained}.}\label{data_statis}
\end{table}

\section{Impact of Graph Reconstruction Loss }\label{alpha}
\begin{figure*}[htb]
    \centering
    \includegraphics[width=0.5\textwidth]{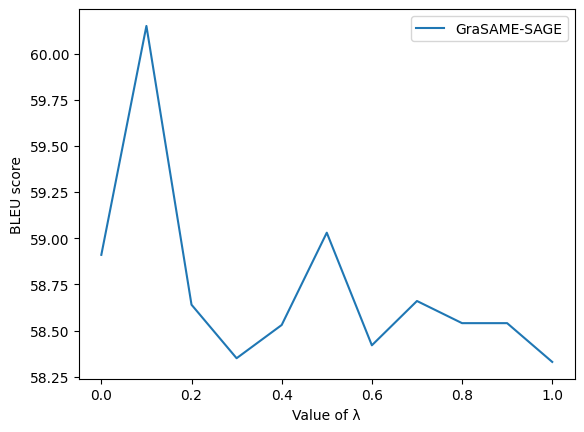}
    \caption{The impact of $\lambda$, experimented with GraSAME-SAGE on the validation set of \textit{WebNLG unconstrained}.}
    \label{figure_alpha}
\end{figure*}

To investigate how graph reconstruction loss affect the performance and to determine the optimal value of $\lambda$ in Equation \ref{eq_10}, we visualize the tuning process in Figure \ref{figure_alpha}. We use GraSAME-SAGE and the validation set of the \textit{WebNLG unconstrained} dataset. The identified best value for $\lambda$ is 0.08.


 

\section{Visualization of the Model Variations}\label{visual_variations}
\begin{figure*}[htb]%
\centering
\scalebox{0.8}{
		\subfloat[][Variation 1: The \colorbox{LightSteelBlue1}{text embedding $x$} is fed into a GNN along with the edge index derived from the hierarchical graph structure. This process generates a \colorbox{Moccasin}{graph embedding $\tilde{x}$}, which subsequently induces the $K$ and $V$ vector. The $Q$ vector then interacts with $K$ and $V$ vector to produce a
    graph-aware representation $\tilde{z}$.]{{\includegraphics[width=9cm]{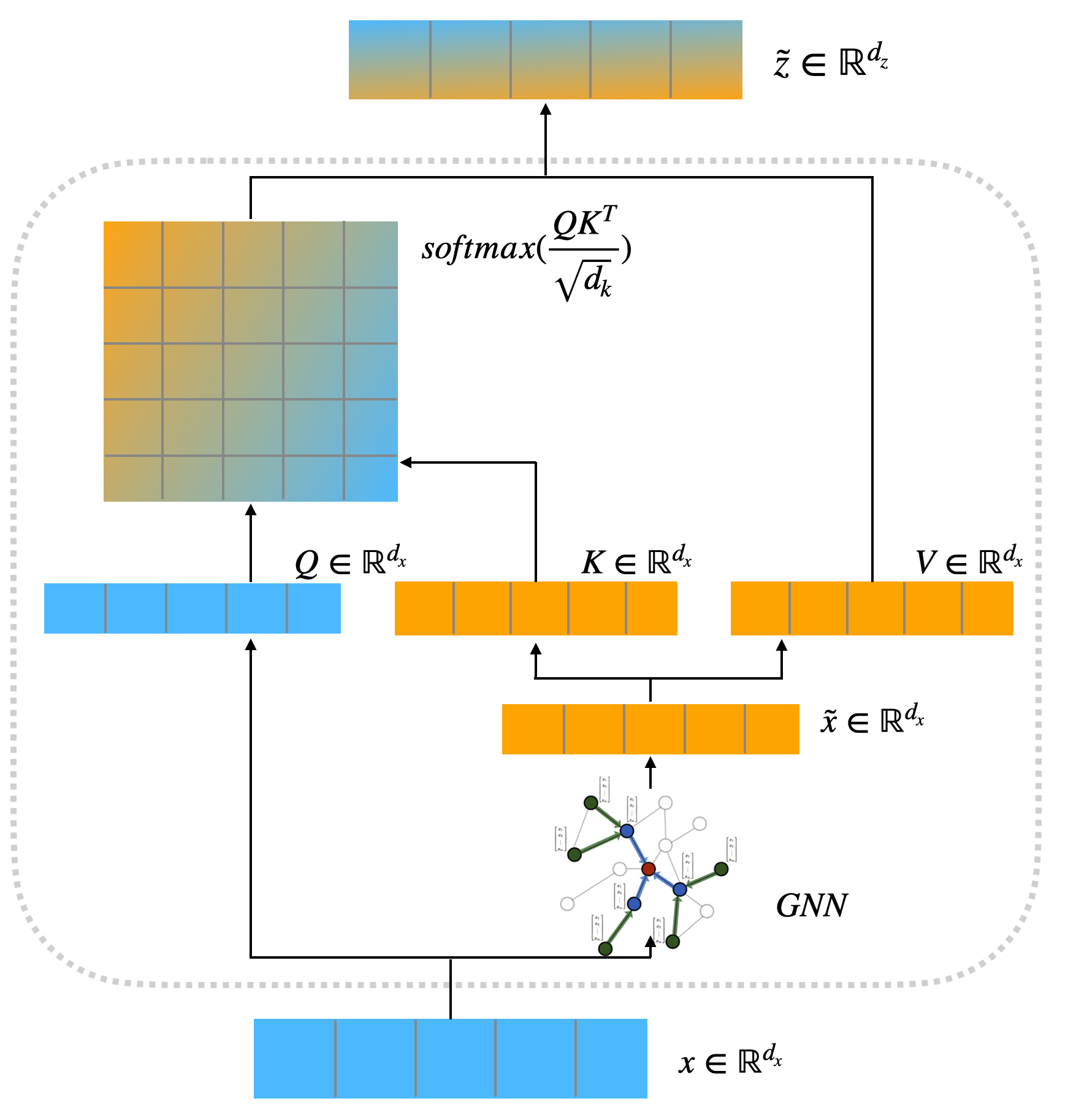} }}%
		\qquad
		\subfloat[][Variation 2: The \colorbox{LightSteelBlue1}{text embedding $x$} is fed into a GNN along with the edge index derived from the hierarchical graph structure. This process generates a \colorbox{Moccasin}{graph embedding $\tilde{x}$}, which subsequently induces the $Q$, $K$ and $V$ vector.]{{\includegraphics[width=9cm]{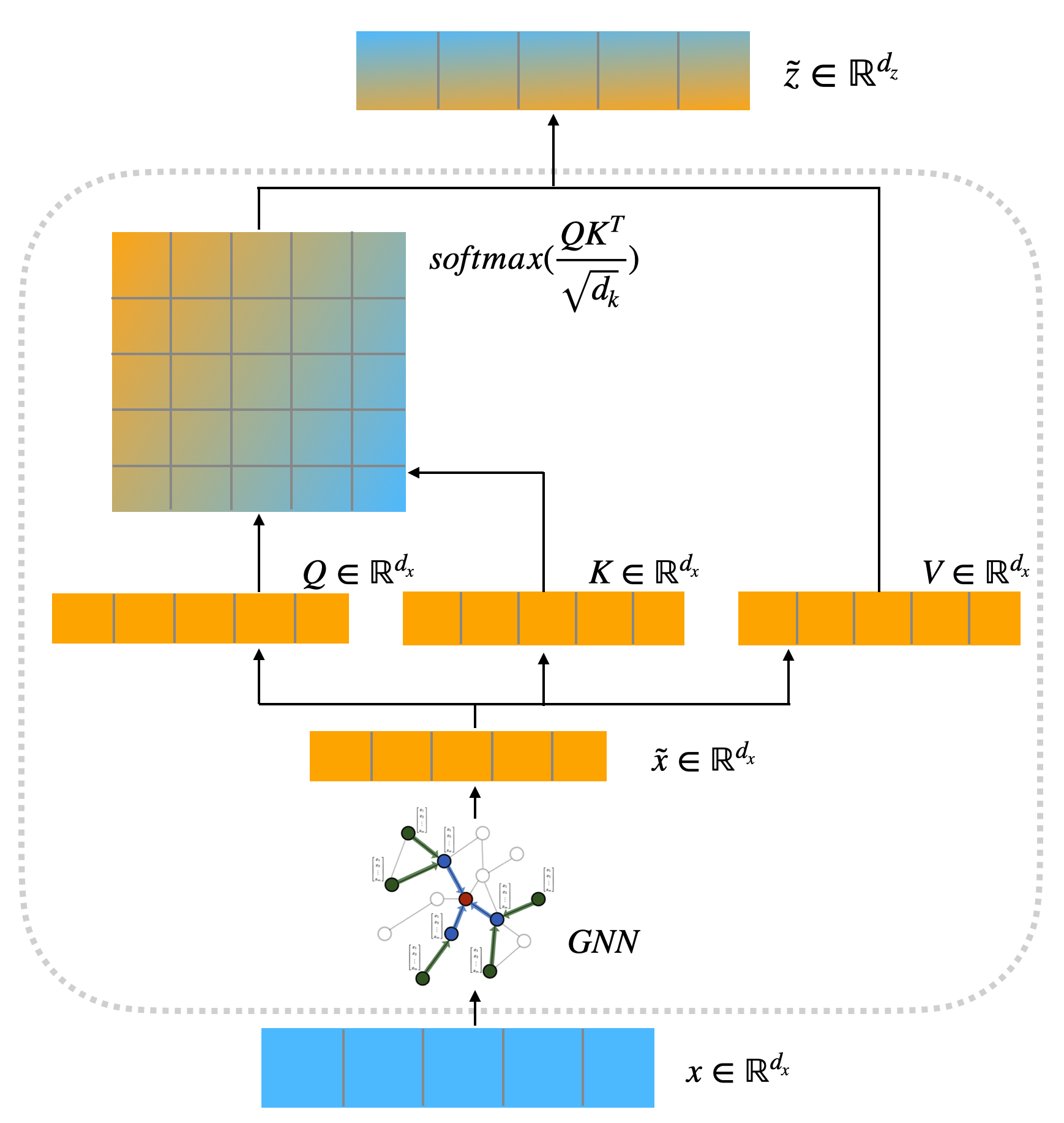} }}}
		\centering\caption{Visualization of two variations of GraSAME.}\label{variation}
	\end{figure*}

To provide a clearer understanding of the model variations, we visualize the internal structure of the self-attention layer for the two model variations in Figure \ref{variation}.

\end{document}